\documentclass[letterpaper, 10 pt, conference]{ieeeconf}  

\IEEEoverridecommandlockouts                              


\usepackage{graphics} 
\usepackage{epsfig} 
\usepackage{mathptmx}  
\usepackage{amsmath}
\usepackage{xcolor}
\usepackage{amssymb}
\usepackage{subcaption}
\usepackage{multicol}
\usepackage{hyperref}
\usepackage{bm}
\usepackage{float}
\usepackage{stfloats}
\usepackage{multicol}
\usepackage[amssymb]{SIunits}

\newtheorem{definition}{Definition}
\newtheorem{remark}{Remark}

\usepackage{cite}

\newcommand\norm[1]{\Big | | #1 | \Big |}
\newcommand{\ith}[1]{$#1^{\text{th}}$}

\usepackage{acro}

\DeclareAcronym{APF}{short = APF, long = Artificial Potential Field}
\DeclareAcronym{CBF}{short = CBF, long = Control Barrier Function}
\DeclareAcronym{CoM}{short = CoM, long = Center of Mass}
\DeclareAcronym{MPC}{short = MPC, long = Model Predictive Controller}

\DeclareMathOperator{\diag}{diag}

\title{\LARGE \bf Reactive Robot-Centric Safety for Autonomous Navigation\\in Constrained and Dynamic Environments}

\author{Viswa Narayanan Sankaranarayanan$^{*}$, Vignesh K. Viswanathan, Akshit Saradagi, \\
Sumeet Satpute, and George Nikolakopoulos
\thanks{*Corresponding author (vissan@ltu.se)}
\thanks{All the authors are with the Robotics and Artificial Intelligence Group of the Department of Computer Science, Electrical and Space Engineering at Lule\aa \ University of Technology, Sweden. This work has been funded by the European Union's Horizon Europe Research and Innovation Program, under the Grant Agreement No. 101119774 SPEAR.}
\thanks{This manuscript is a preprint and is currently under review.}}
\begin{document}
\maketitle

\begin{abstract}
In this work, we address the problem of ensuring real-time safety in autonomous robot navigation, in spatially constrained dynamic environments, by utilizing only onboard sensors. We present a real-time control architecture that integrates a 3D LIDAR perception-based composite \ac{CBF}-based safety filter directly into the autonomy pipeline. The proposed perception-driven framework enforces collision avoidance constraints dynamically from onboard point cloud data, thus allowing a large number of constraints to be handled at the control frequency, while remaining minimally invasive to nominal task execution. The safety region is defined as an ellipsoid in the body-frame, consistent with the geometry of the platform, which induces time-varying constraints in the world frame as the robot rotates; this effect is handled through a dedicated formulation of time-varying \ac{CBF} for each LIDAR point. We validate the system through multiple field experiments in underground environments by utilizing a quadruped platform performing a visual inspection task, demonstrating reliable operation in the presence of dynamic obstacles, unsafe high-level references, abrupt localization anomalies, and while traversing through narrow corridors.
\end{abstract}
%
\section{Introduction} \label{sec:intro}
\begin{figure}[htbp]
    \centering
    \includegraphics[width=0.9\linewidth]{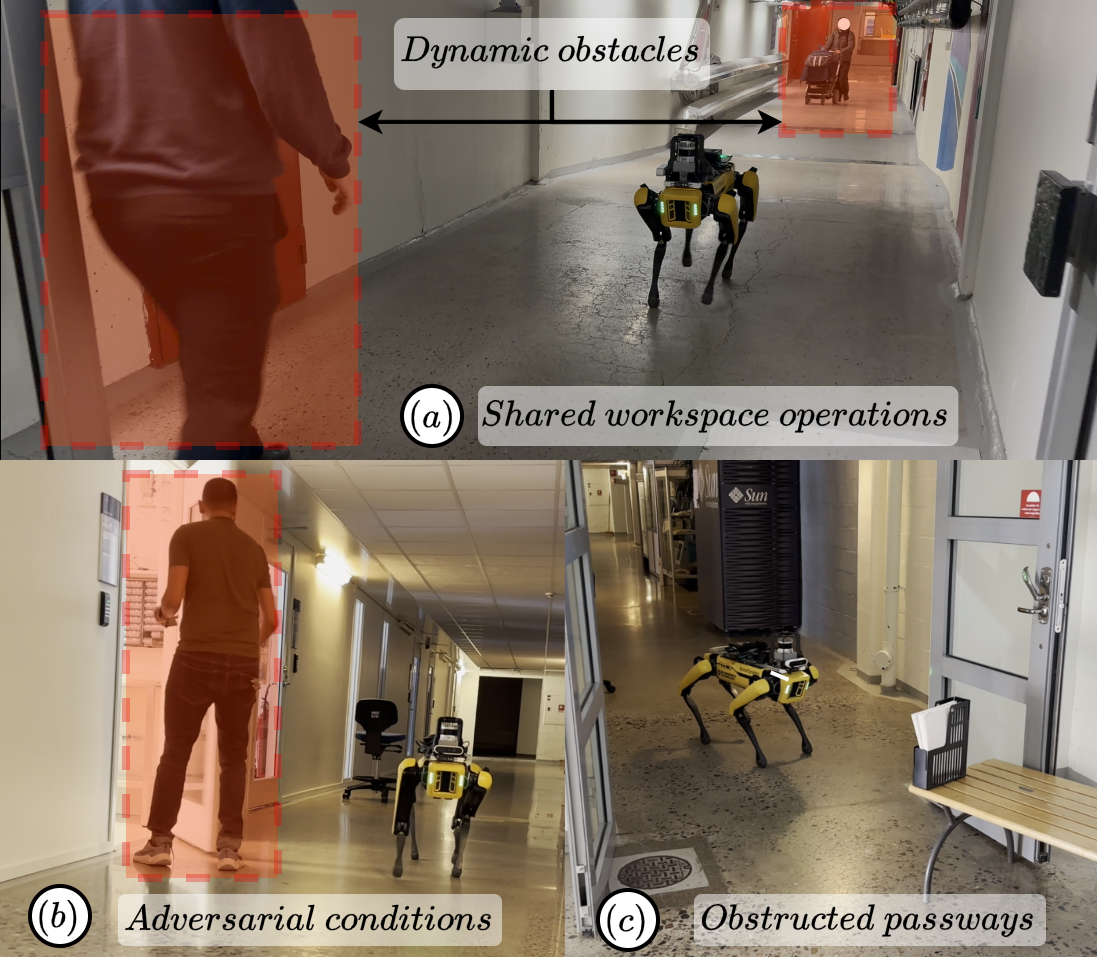}
    \caption{A visual illustration of real-world operating conditions demonstrating dynamic and uncertain scenes faced during autonomous missions.}
    \label{fig:sentry_scene}
\end{figure}

Autonomous robots are increasingly being deployed for inspection, maintenance, and exploration in spatially cluttered environments, such as subterranean mines, warehouses, forests, and industrial facilities. By reducing human involvement in hazardous environments, these systems improve operational safety and efficiency. Despite the advances in onboard autonomy, ensuring robot safety is a critical challenge in autonomous navigation missions. In particular, for inspection and exploration missions, robots must operate in proximity to the infrastructure, often in a partially known, dynamic, or adversarial environment. 
Figure~\ref{fig:sentry_scene} illustrates such representative operating conditions encountered during autonomous missions. In Figure~\ref{fig:sentry_scene}(a), the robot operates in an active, shared workspace with pedestrians, requiring safe co-navigation under limited lateral clearance. Figure~\ref{fig:sentry_scene}(b) depicts more adversarial scenarios in which unanticipated human motion and other dynamic changes challenge the nominal operation and safety. In Figure~\ref{fig:sentry_scene}(c), environmental artifacts left by other users (e.g., furniture and equipment near doorways) reduce free space and create further narrow passageways, motivating responsive safety enforcement to maintain smooth tracking while respecting obstacle-avoidance margins.

Beyond external disturbances, robot safety is further challenged by internal uncertainties, arising from perception and state estimation pipelines. Specifically, in geometrically repetitive environments (e.g., long corridors with symmetric layouts), the localization problem could become ill-conditioned due to geometric degeneracy and can exhibit perceptual aliasing~\cite{lajoie2019modeling}. These effects may manifest as abrupt odometry drift or incorrect loop closures, directly increasing the likelihood of safety constraint violations~\cite{zhang2016degeneracy}. Collectively, these challenges motivate an inherent onboard, sensor-driven safety layer that enforces safety even when mapping, localization, or planning outputs become stale, inconsistent or exhibit slower response to rapid, dynamic changes.

Recent advancements in instantaneous point-cloud-based reactive navigation methods enable robots to operate safely while executing their tasks by leveraging real-time geometric measurements to enforce safety constraints directly at the control level, without fully relying on a high-level planner. While this approach improves robustness in unknown or dynamic environments, it also introduces significant challenges: safety must be guaranteed from dense point cloud streams under resource-constrained onboard hardware.

While \ac{MPC}s are commonly used for reactive navigation~\cite{marzat2017reactive}, incorporating dense point cloud–based constraints often leads to nonlinear optimization problems that are computationally demanding~\cite{falanga2018pampc}. Compared to Artificial Potential Field-based methods, \ac{CBF}-based safety filtering provides smoother and less oscillatory behavior, while remaining minimally invasive, modifying the nominal control input only when safety is at risk~\cite{singletary2021comparative}. CBFs offer a computationally efficient mechanism for enforcing forward invariance of safety sets through affine constraints in a quadratic program, making them well-suited for real-time onboard deployment~\cite{ames2019control}.

In the literature, constraints for reactive navigation \cite{marzat2017reactive, falanga2018pampc, singletary2021comparative} are typically designed from an exterior perspective that specifies regions the robot must avoid. Such formulations do not explicitly account for the robot body's geometry. Since robot bodies are not necessarily isotropic, using a distance-based constraint for every observed point results in spherical constraints that can be overly conservative. Enabling robots to traverse narrow passages, therefore, requires an axis-aligned ellipsoidal safety envelope defined in the body frame (cf. Figure~\ref{fig:frames}), consistent with the platform geometry and less conservative than spherical approximations. This formulation reflects a robot-centric safety perspective, where safety constraints are defined relative to the robot’s body geometry and orientation rather than solely from the obstacle locations. This representation preserves the directional clearance margins relative to the body frame. However, the resulting point-based constraints become functions of the robot’s orientation, leading to a time-varying safety envelope even in the presence of static obstacles. Such orientation-dependent constraints can be naturally incorporated within a control barrier function framework to handle time-varying safety requirements as in~\cite{Lindermann2019Control}. In addition, composite CBF formulations~\cite{molnar2023composing} provide a systematic way to combine multiple point-wise safety constraints into a single smooth barrier function~\cite{harms2025safe}, allowing the simultaneous imposition of all constraints at the control frequency. In addition, composite CBF frameworks also support time-varying CBF structures~\cite{safari2024time}, enabling the representation of body-centered safety envelopes.

While CBF-based safety filtering has shown promising results, experimental demonstrations of perception-driven safety filtering operating at control frequency within an autonomy framework in spatially constrained real-world environments remain limited. In particular, its deployment using geometry-consistent body-frame safety envelopes with dense onboard point cloud data, and under dynamic or practically challenging conditions, has not been systematically studied. Motivated by these observations, we present the main novel contributions of this work.

\textbf{Contributions:} This article presents a real-time safety-critical control architecture for robots operating in spatially constrained environments. The proposed system integrates a perception-driven composite \ac{CBF}-based safety filter, directly within the autonomy pipeline, enabling collision avoidance constraints to be enforced at control frequency from dense onboard point cloud data. A key design element is the formulation of the safety region as an axis-aligned ellipsoid in the body frame that reflects the platform geometry, allowing less conservative and direction-aware clearance compared to conventional distance-based constraints. This representation naturally induces orientation-dependent time-varying constraints in the world frame during rotational motion, for which we develop a corresponding time-varying \ac{CBF} formulation for each point within the point cloud. The effectiveness of the architecture is validated through extensive field experiments in underground environments on a quadruped platform, including narrow corridor traversal, handling dynamic obstacles, and robustness to high-level anomalies such as unsafe references and abrupt odometry jumps.



\section{Problem Formulation} \label{sec:prob_form}
In this Section, we define the class of robots under consideration in this work,  identify the safety requirements for autonomous navigation in a spatially constrained dynamic environment, and explicitly formulate the problem statement addressed in this work. 

\subsection{System Modeling} \label{subsec:sys_model}
We consider a broad class of robot platforms whose dynamics can be written in the control-affine form:
\begin{align}
\dot{q} = f_q(q) + g_q(q)\tau, \label{eq:dyn_model}
\end{align}
where $q \in \mathbb{R}^n$ is the platform-dependent system state and 
 $\tau \in \mathbb{R}^m$ is the vector of actuation inputs. $f_q(q)$ and $g_q(q)$ denote the drift and control vector fields respectively. The position of the robot in the world frame, part of the system state $q$, is denoted as $p_r^W$.

\subsection{Robot-centric Safety Constraints from Local LIDAR Scans} \label{subsec:safe_cons}
    

\begin{figure}[ht]
    \centering
    \includegraphics[width=\linewidth]{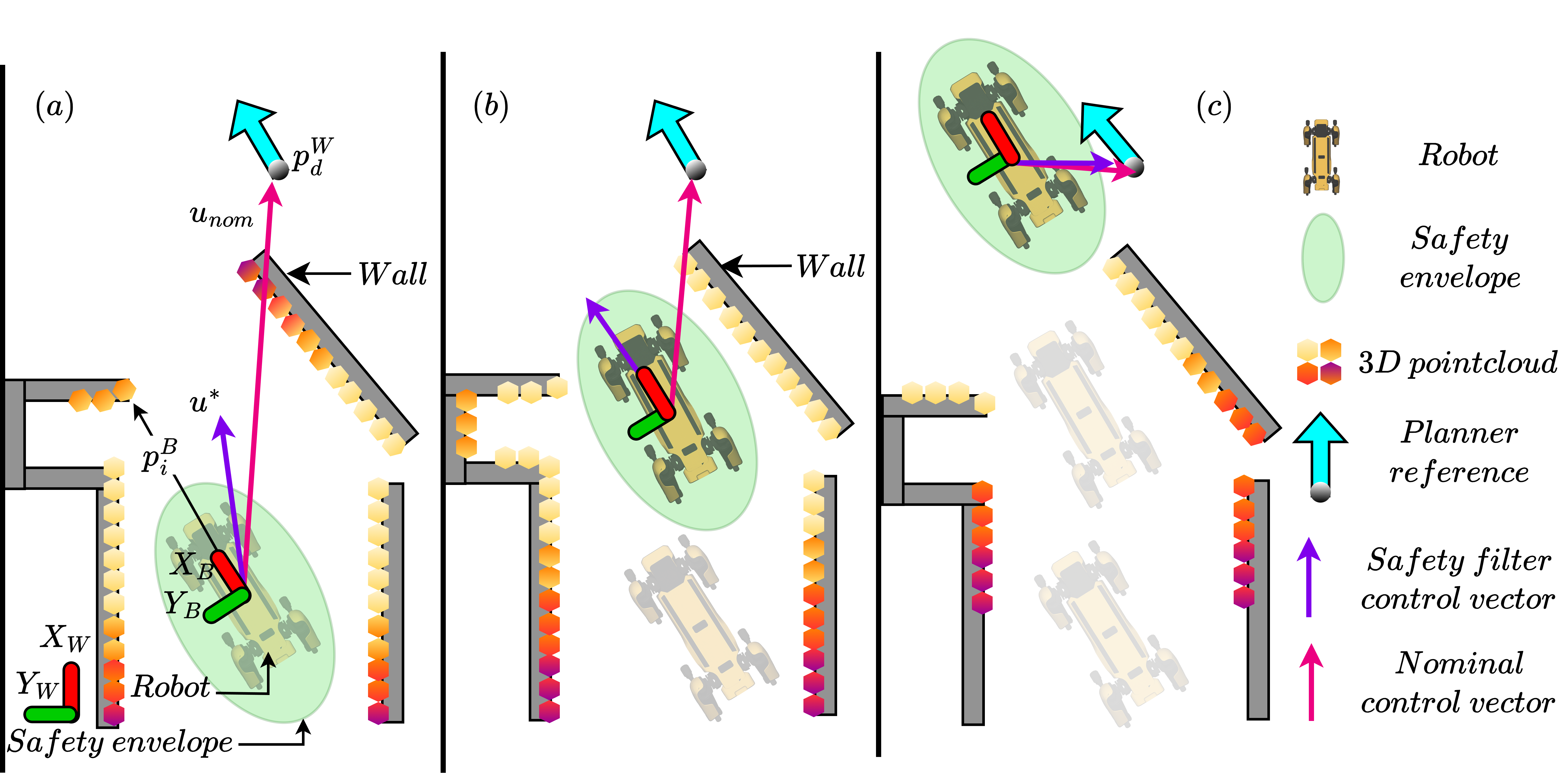}
    \caption{An illustration of the robot-centric safety. Body-centric safety envelope enables the robot to keep away from the locally scanned LIDAR points while navigating through spatial constraints to reach the goal.}
    \label{fig:frames}
\end{figure}

To derive collision avoidance constraints from onboard LIDAR perception, we define an inertial world frame $\mathbf{X_W} - \mathbf{Y_W} - \mathbf{Z_W}$ and a body-fixed frame $\mathbf{X_B} - \mathbf{Y_B} - \mathbf{Z_B}$ attached to the robot as shown in Figure \ref{fig:frames}. At any instant $t$, the obstacles in the neighborhood of the robot appear in the current local LIDAR scan as points, and are represented in the body frame as $p^B_{i}(t) \triangleq [x^B_i(t), y^B_i(t), z^B_i(t)]^\top$, where $i \in \{1, 2, ..., N\}$ and with $N$ the number of obstacle points obtained from a single scan. 

In this work, we consider the general scenario of a robot with an anisotropic footprint (Quadruped robots, aerial manipulators, and Humanoids being some examples). Such a robot must maintain a different safety distance from each of the LIDAR points along different axes. To simplify the derivation of safety constraints with respect to LIDAR points, we approximate the safe region around the robot as an ellipsoid around the robot's actual footprint. Given the safety region, a robot-centric safety necessitates that the ellipsoid must be kept free of any of the points from the current LIDAR scan. Therefore, the robot-centric safety constraints in the body frame of the robot are modeled as:
\begin{align}
    \left (\frac{x^B_i}{a_x} \right )^2 + \left (\frac{y^B_i}{a_y} \right )^2 + \left (\frac{z^B_i}{a_z} \right )^2 \geq 1 \;\; \forall \; i  \in \{1, 2, ..., N\} \label{eq:point_constraint_body}
\end{align}
where $a_x, a_y, a_z > 0 $ are the semi-axes lengths of the safety ellipsoid, chosen based on the robot geometry, such that the robot is completely enclosed by the ellipsoid, as shown in Figure \ref{fig:frames}. With respect to the position of the robot in the world frame ($p_r^W$), the safety constraints can be written as: 
\begin{align}
    (p^W_i - p^W_r)^\top Q_W (p^W_i - p^W_r) \geq 1 \label{eq:point_constraint},
\end{align}
where $Q_W = R^W_B(t) \diag \left (\frac{1}{a^2_x}, \frac{1}{a^2_y}, \frac{1}{a^2_z} \right ) (R^W_B(t))^\top$ is the world frame shape matrix.

\begin{remark}[Planar case]
 Although the safety envelope in~\eqref{eq:point_constraint} is formulated as a 3D ellipsoid for generality, the same formulation directly applies to planar platforms by restricting the constraint to the $(x,y)$ subspace, resulting in a 2D elliptical safety region. In the experimental validation on the quadruped platform presented in this work (Section \ref{sec:exp_val}), we employ this planar specialization.
\end{remark}


\noindent \textbf{Problem Statement:} Given the robot dynamics in~\eqref{eq:dyn_model}, a robot-centric safety ellipsoid for a robot with anisotropic footprint and the obstacle points, obtained from onboard LIDAR perception, develop a control architecture that ensures that the robot-centric safety constraints in~\eqref{eq:point_constraint} are enforced in real time (using limited onboard computational resources), allowing robots to navigate and accomplish tasks in constrained dynamic environments.

\section{Proposed Control Architecture} \label{sec:prop_meth}

\begin{figure}[htbp]
    \centering
    \includegraphics[width=0.9\linewidth]{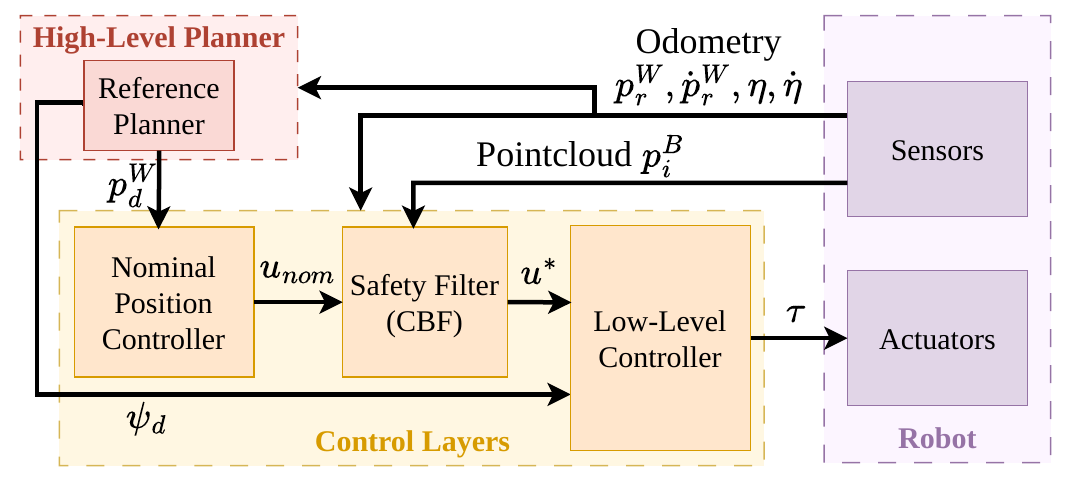}
    \caption{
    The multi-loop control architecture proposed in this article for enabling safe navigation during autonomous missions.}
    \label{fig:control_arch}
\end{figure}

This Section presents the overall multi-loop control architecture, shown in Figure \ref{fig:control_arch}, as the solution to the problem defined in Section~\ref{sec:prob_form}. The proposed novel multi-loop architecture consists of a nominal position controller, a CBF-based safety filter, a low-level controller, and the provision to fit in a high-level task planner (to provide references for the nominal controller). The objective of the CBF-based safety filter is to enforce the local LIDAR scan-based robot-centric safety constraints presented in~\eqref{eq:point_constraint} over the potentially-unsafe nominal control inputs. Since these safety constraints require the local LIDAR points to remain outside the safe ellipsoid around the robot, they result in nonconvex feasible sets in the state space. Enforcing such nonconvex state constraints, one for each LIDAR point within the conventional \ac{MPC} framework, results in large nonlinear and nonconvex optimization problems. The existing iterative solvers for such nonlinear and nonconvex problems are computationally heavy and cannot be solved onboard resource-constrained robot platforms at high control frequencies necessary for reactive navigation in dynamic environments.

In this article, the primary objective is to design a computationally efficient methodology to enforce the robot-centric safety constraints in real-time. To achieve the objective, we recast the constraints in Equation \ref{eq:point_constraint}, as Control Barrier Functions (\ac{CBF}s) and employ a \ac{CBF}-based safety filter to derive safe control actions. The CBF-based safety filter takes in a nominal velocity input generated by the nominal position controller and generates a minimally invasive filtered velocity input that satisfies all the safety constraints. This filtered velocity input is then tracked by an appropriate platform-dependent low-level controller, as depicted in the multi-loop architecture in Figure \ref{fig:control_arch}.

In this work, we consider a go-to-goal proportional controller to generate a nominal velocity input $u_{nom} = K_p  (p^W_d - p^W_r)$, where $K_p \in \mathbb{R}^{3 \times 3}$ is a 
diagonal gain matrix and $p^W_d \in \mathbb{R}^3$ is the desired position. The safety filter modifies $u_{nom}$ to produce $u^* \in \mathbb{R}^3$, which is passed to the low-level controller to generate $\tau$, and mapped to the actuator commands.

\begin{figure}[ht]
    \centering
    \includegraphics[width=0.9\linewidth]{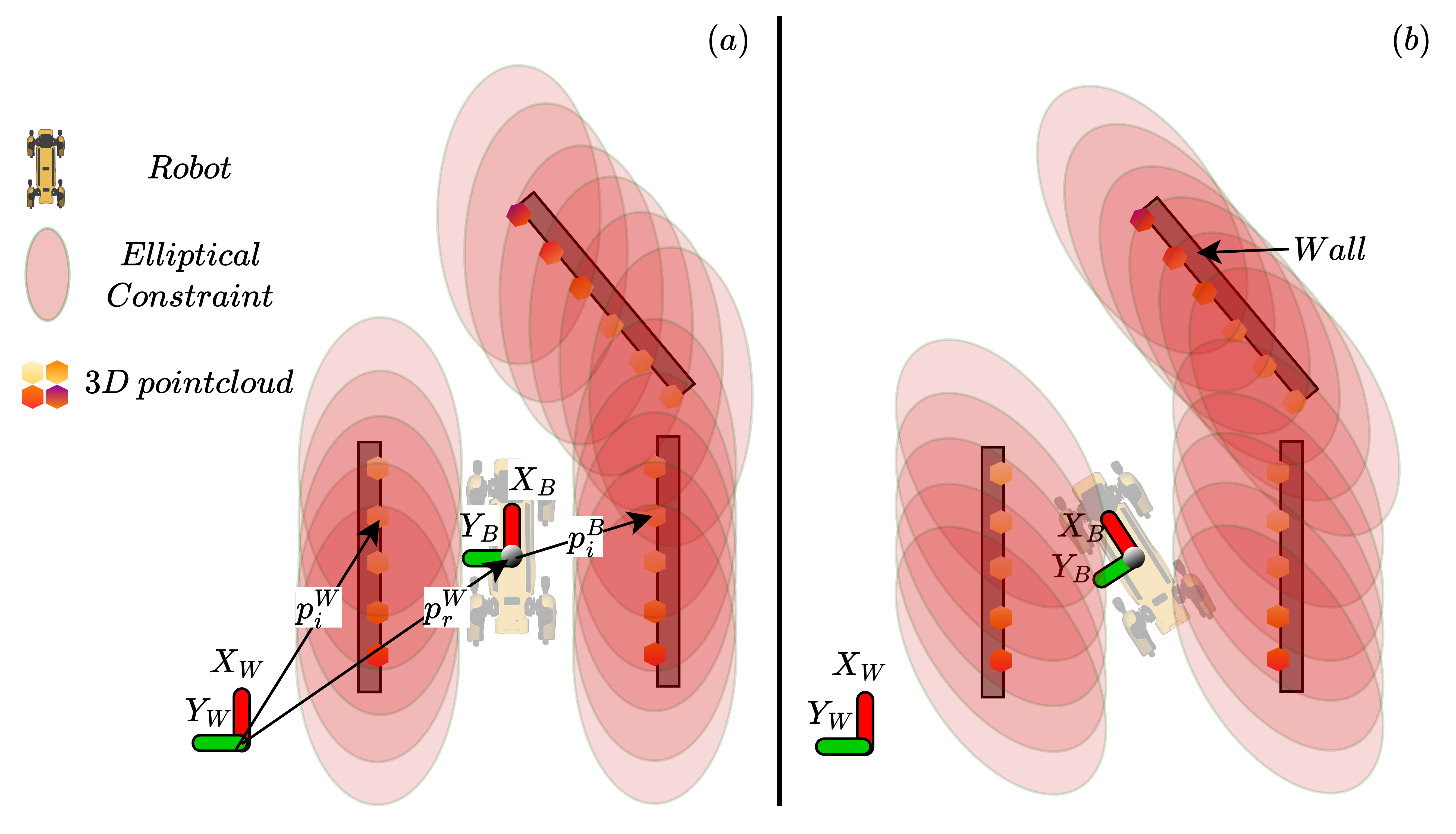}
    \caption{A dual interpretation of the ellipsoidal constraints in the world frame, centered on each of the LIDAR points, whose axes are aligned with the robot's bodyframe. For the same positions of the robot and LIDAR points between (a) and (b), the direction of the ellipsoids changes with respect to the robot's orientation $R^W_B (t)$, making the constraints time-varying when the robot rotates.}
    \label{fig:concept_dual}
\end{figure}

Although the obstacles are assumed to be static, the constraints depend on the rotation matrix, which is time-varying and not explicitly modeled in the translational dynamics of the robot (cf. Figure \ref{fig:concept_dual}). We therefore begin with the introduction of time-varying \ac{CBF} in Subsection \ref{subsec:tvcbf}, and construct our robot-centric safety constraints in Subsection~\ref{subsec:cbf_design}. Furthermore, to impose a large number of LIDAR point-based constraints simultaneously in a scalable manner, we adopt a composite \ac{CBF} approach, which will be introduced in Subsection~\ref{subsec:comp_cbf}. Finally, we construct the composite CBF using the individual \ac{CBF}s and form the CBF-safety filter (a quadratic program). For brevity, the low-level control layer is abstracted and assumed to track the filtered velocity command, since its design depends on the specific robot platform.

    \subsection{Time-varying Control Barrier Function} \label{subsec:tvcbf}
Consider the control-affine system
$
    \dot{p} = f(p) + g(p)u,
$
where $p \in \mathcal{P} \subset \mathbb{R}^n$, 
$u \in \mathcal{U} \subset \mathbb{R}^m$, 
and $f$ and $g$ are locally Lipschitz continuous functions. Let the time-varying safe set be defined as: $
    \mathcal{S}(t) := \{ p \in \mathcal{P} \mid h(p,t) \ge 0 \},$ where $h :  \mathcal{D}(t) \subset \mathcal{P} \times \mathbb{R}_{\ge 0} \rightarrow \mathbb{R}$ is continuously differentiable in both $p$ and $t$. The set $\mathcal{S}(t)$ is rendered safe if the control input $u$ ensures forward invariance, i.e., $
p(t_0) \in \mathcal{S}(t_0) 
\;\Rightarrow\; 
p(t) \in \mathcal{S}(t), 
\quad \forall t \ge t_0.
$

\begin{definition}[Time-Varying \ac{CBF}] \label{def:tvcbf}
A continuously differentiable function $h(p,t)$ is said to be a valid time-varying \ac{CBF} for the system $\dot{p}=f(p)+g(p)u$ if there exists a locally Lipschitz continuous extended class-$\mathcal{K}$ function $\alpha(\cdot)$ such that, for all $(p,t) \in \mathcal{P} \times [0,\infty)$,
 \begin{equation}
    \sup _{u \in \mathcal{U}}  \left(\frac{\partial h(p, t)}{\partial p} ^\top f(p)+g(p)u+\frac{\partial h(p, t)}{\partial t}\right) \geq -\alpha(h(p, t)).
    \label{eq:tvcbf_def}
\end{equation}
\end{definition}

\begin{remark}
The criterion in~\eqref{eq:tvcbf_def} guaranties the forward invariance and the asymptotic stability of the safe set $\mathcal{S}(t)$, through the relationships $\dot{h}(p,t) \geq 0$ at the boundary of the safe set $\partial \mathcal{S}$, and $\dot{h}(p,t) > 0$ outside the safe set, i.e., $\mathcal{D}(t)\setminus \mathcal{S}(t))$.
\end{remark}

    \subsection{CBFs corresponding to robot-centric safety constraints} \label{subsec:cbf_design}

For the \ith{i} point's safety constraint, we form the following time-varying CBF.
\begin{align}
    h_{Wi}(p^W_r, t) &= (p_i^W - p^W_r)^\top Q_W(t)(p_i^W - p^W_r) - 1.
\end{align}
The associated safe set is given by the zero superlevel set:
\begin{align}
    \mathcal{S}_{Wi} = \left\{\, p^W_r \in \mathbb{R}^3 \;\middle|\; h_{Wi}(p^W_r,t)\ge 0 \,\right\}.
\end{align}

Since velocity is the control input in this layer, we base our CBF design on the first-order kinematic model, $\dot{p}^W_r = u^W$, where $u \in \mathbb{R}^{3}$ is the velocity input. Comparing this model with the control affine dynamics in Definition \ref{def:tvcbf}, we have $f(p) = 0$, $g(p) = \mathbf{I}$. So, the partial derivatives of $h_i$ are as follows.
\begin{align}
    \begin{bmatrix}
         \frac{\partial h_{Wi}}{\partial p^W_r} \\
         \frac{\partial h_{Wi}}{\partial t} 
    \end{bmatrix}   & = \begin{bmatrix}
         -2 Q_W(t) (p^W_i - p^W_r) \\ (p_i^W - p^W_r)^\top \dot{Q}_W(t)(p_i^W - p^W_r) 
    \end{bmatrix}   
\end{align}
For every one of the $N$ points in the current local LIDAR scan, there exists a corresponding CBF. For dense point clouds, this leads to scalability challenges associated with implementing the CBF filter, as a large quadratic program has to be solved at a high frequency. In this context, the concept of a Composite Control Barrier function, recently proposed in the literature, is extremely beneficial.  

    \subsection{Composite Control Barrier Function} \label{subsec:comp_cbf}
    In this subsection, we combine the individual barrier functions defined in the previous section to construct a single smooth composite function using a log-sum-exp (soft-min) operator~\cite{molnar2023composing}. We first formally define the soft-min composite barrier function.

\begin{definition}[Soft-Min Composite Barrier Function~\cite{molnar2023composing, harms2025safe}] \label{def:softmin_composite}
    Let $\{h_i : \mathcal{P} \rightarrow \mathbb{R}\}_{i=1}^N$ be continuously differentiable functions and define the individual safe sets
    \begin{equation}
        \mathcal{S}_i := \{ p \in \mathcal{P} \mid h_i(p) \ge 0 \}.
    \end{equation}
    For fixed parameters $\kappa > 0$ and $\gamma > 0$, the soft-min composite barrier associated with $\{h_i\}_{i=1}^N$ is the continuously differentiable function $H : \mathcal{P} \rightarrow \mathbb{R}$ defined as
    \begin{equation}    
        H(p) = -\frac{\gamma}{\kappa} \log \left( \sum_{i=1}^{N} \exp \left( -\kappa  \tanh\!\left(\frac{h_i(p)}{\gamma}\right)  \right)  \right). \label{eq:softmin_definition}
    \end{equation}
    The composite safe set is defined as
    \begin{equation}
        \mathcal{S}_H := \{ p \in \mathcal{P} \mid H(p) \ge 0 \}.
    \end{equation}
\end{definition}

\begin{remark}[Set Inclusion Property] \label{rem:softmin_inclusion}
    The log-sum-exp soft-min constitutes a smooth under-approximation of the pointwise minimum operator. In particular, for all $p \in \mathcal{P}$, it holds that $H(p) \le \min_i \left ( \tanh\!\left(\frac{h_i(p)}{\gamma}\right) \right )$ . Consequently, $H(p) \ge 0 \Rightarrow h_i(p) \ge 0,\ \forall i = 1,\dots,N$, and therefore $\mathcal{S}_H \subseteq \bigcap_{i=1}^{N} \mathcal{S}_i$. Hence, nonnegativity of the composite barrier provides a sufficient condition for the simultaneous satisfaction of all individual barrier constraints. The log-sum-exp operator provides a smooth approximation of the minimum; similar composite barrier constructions appear in \cite{molnar2023composing}.
\end{remark}
        
\noindent The parameters $\kappa$ and $\gamma$ control the sharpness and saturation level of the soft-min aggregation, respectively. Since the logarithm, exponential, and hyperbolic tangent functions are smooth, $H(p)$ is continuously differentiable whenever each $h_i(p)$ is continuously differentiable.

\noindent
\textbf{Time-varying case:} In the setting of this article, the individual barrier functions are time-varying. To cater to the current setting, we employ the same soft-min construction to define a notion of a time-varying composite barrier function:
\begin{align}
    H(p,t) = -\frac{\gamma}{\kappa} \log \left( \sum_{i=1}^{N} \exp \left( -\kappa \tanh\!\left(\frac{h_i(p,t)}{\gamma}\right) \right) \right).
\end{align}
Since each $h_i(p,t)$ is continuously differentiable in $(p,t)$, the composite function $H(p,t)$ is also continuously differentiable. For each fixed $t \ge 0$, the soft-min satisfies $H(p,t) \le \min_i \left ( \tanh\!\left(\frac{h_i(p,t)}{\gamma}\right) \right )$. Consequently, $H(p,t) \ge 0 \Rightarrow h_i(p,t) \ge 0,\ \forall i$, and therefore $\mathcal{S}_H(t) \subseteq \bigcap_{i=1}^{N} \mathcal{S}_i(t)$. We therefore treat $H(p,t)$ as a time-varying CBF and enforce the condition of Definition~\ref{def:tvcbf} directly on $H(p,t)$, i.e.,
\begin{align}
    \frac{\partial H(p,t)}{\partial p}^\top (f(p)+g(p)u) + \frac{\partial H(p,t)}{\partial t} \ge -\alpha(H(p,t)).
\end{align}



Now, we form one aggregated composite CBF by combining the $N$ individual CBFs $h^W_i$ (in subsection \ref{subsec:cbf_design}) corresponding to local LIDAR points, as shown next.
\begin{align}
    H_W(p_r^W,t) &=  -\frac{\gamma}{\kappa} \log\!\left(  \sum_{i=1}^{N}  \exp\!\left(  -\kappa  \tanh\!\left(\frac{h_{Wi}}{\gamma}\right)  \right)  \right).    \label{eq:HW_composite_prW}
\end{align}
The gradients of $H_W$ are:
\begin{align}
    \frac{\partial H_W}{\partial p_r^W} &= \sum_{i=1}^{N} \lambda_i(p_r^W,t)\, \frac{1} {\cosh^2\! \left ( \frac{h_{Wi}}{\gamma} \right )} \frac{\partial h_{Wi}}{\partial p_r^W}, \\
    \frac{\partial H_W}{\partial t} &= \sum_{i=1}^{N} \lambda_i(p_r^W,t)\, \frac{1}{\cosh^2\! \left ( \frac{h_{Wi}}{\gamma} \right )} \frac{\partial h_{Wi}}{\partial t}, 
\end{align}
where 
\begin{align}
    \lambda_i  &= \frac{ \exp\! \left ( -\kappa \tanh\!\left ( \frac{h_{Wi}} {\gamma} \right ) \right ) } { \sum_{j=1}^{N} \exp\! \left ( -\kappa \tanh\! \left ( \frac{h_{Wj}}{\gamma} \right ) \right )}, \qquad \sum_{i=1}^{N}\lambda_i=1.
\end{align}

Now, we obtain the filtered control velocity $u^*$ using the following quadratic program that imposes the safety constraint \eqref{eq:HW_composite_prW} on the nominal control input.
\begin{align}
    u^* (p^W_r, t) &= \underset{u \in \mathcal{U}}{\text{argmin}} \norm{u - u_{nom}(p^W_r, t)}, \nonumber \\ 
    \text{s.t.} &: \left(\frac{\partial H_W}{\partial p_r^W}\right)^\top u \geq -\alpha(H_W) - \frac{\partial H_W}{\partial t}. \label{eq:qp}
\end{align}
For a given $p^W_r$, the constraint \eqref{eq:qp} is linear in $u$, and therefore the optimization problem is a standard quadratic program, which can be solved efficiently at high frequencies.

\section{Experimental Evaluation} \label{sec:exp_val}

We validate the proposed controller on an inspection planning mission. The high-level planner and the experimental setup used in the mission are presented in Subsections \ref{subsec:insp_plan} and \ref{subsec:exp_set}, respectively. The safety filter's effectiveness in ensuring local robot safety is validated in three scenarios: Scenario 1 demonstrates the need for body-centric constraints in narrow corridor traversal; Scenarios 2 and 3 present the performance of the safety filter in the presence of drastic odometry drift, and dynamic obstacles, respectively.
\subsection{Inspection Planning} \label{subsec:insp_plan}
We use the reactive local view planner~\cite{viswanathan2024surface} as the high-level reference planner. The planner operates over instantaneous 3D point cloud measurements to generate the \ith{(k+1)} reference view pose $[(p^W_d(k+1))^\top ~ \psi(k+1)]^\top$, satisfying the desired viewing constraints. At each planning instance, the nearest 3D point on the locally observed surface $p^W_{nn} (k)$ is obtained. From this, the reference unit vectors capturing the ego orientation of the robot are determined. Based on the viewing constraints, $d_{view},\gamma_H,\gamma_V$, the next view-pose is computed with respect to the robot's current position. The update rule for the planner is given by,
\begin{equation}\label{eqn:flip}
  p^W_d(k+1) = p^W_r(k) + {\nu}_x d_{insp} + \nu_y d_{hov} + \nu_z d_{vov}
\end{equation}
where,\begin{equation*}\label{eqn:unit_vec}
\begin{aligned}
    \nu_x(k) &= \frac{p^W_{nn}(k) - \hat{p}^W_r(k)}{\norm{p^W_{nn}(k) - \hat{p}^W_r(k)}} \\
    \nu_y^k &= \nu_{up}\times  \nu_x^k, ~
    \nu_z^k = \nu_x^k \times \nu_y^k \\
    d_{hov} &= 2\tan(\frac{\alpha}{2})\norm{p^W_{nn}(k) - \hat{p}^W_r(k)} \\
    & - 2\tan(\frac{\alpha}{2})\norm{p^W_{nn}(k) - \hat{p}^W_r(k)} \gamma_H \\
    d_{vov} &= 2\tan(\frac{\beta}{2})\norm{p^W_{nn}(k) - \hat{p}^W_r(k)} \\& - 2\tan(\frac{\beta}{2})\norm{p^W_{nn}(k) - \hat{p}^W_r(k)} \gamma_V \\
    d_{insp} &= ||{p}^W_{nn}(k) - \hat{p}^W_{r}(k) || - d_{view}
\end{aligned}
\end{equation*}
\noindent such that, ${\nu_x},{\nu_y},{\nu_z} \in \mathbb{R}^3$ are the unit view vectors of the robot with $\nu_{up} = [0,0,1] $. The desired yaw is given by,
 \begin{equation}\label{eqn:flip_yaw}
     \psi({k+1}) =\arctan(\nu_x^{k+1}(1),\nu_x^{k+1}(0))
 \end{equation}

\begin{figure}[htbp]
    \centering
    \includegraphics[width=0.8\linewidth]{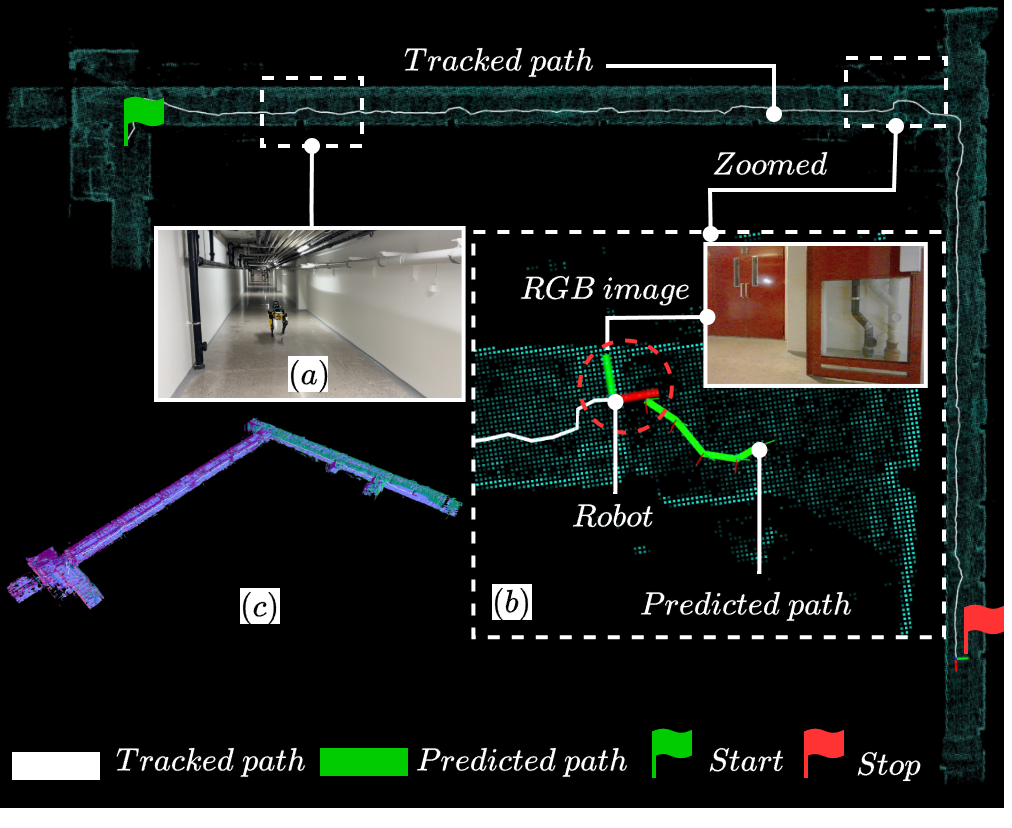}
    \caption{Visual representation of the inspection planner during autonomous deployment in underground corridors.}
    \label{fig:FLIP_overall}
\end{figure}

The predicted inspection path $\bm{\Pi}^k = \{[(p^W_d(k+1))^\top ~ \psi(k+1)]^\top\}^N_{m=1}$ is computed over a horizon $N \in \mathbb{R}_+$ by recursively evaluating~\eqref{eqn:flip}-\eqref{eqn:flip_yaw} for the current point-cloud observation. Figure~\ref{fig:FLIP_overall} presents the runtime snapshot of the inspection planner during field deployment. Figure~\ref{fig:FLIP_overall}(a) captures the robot during inspection in underground corridors. Figure~\ref{fig:FLIP_overall}(b) highlights the outcome of the inspection planner as the robot navigates through a narrow doorway and Fig~\ref{fig:FLIP_overall}(c) presents the 3D reconstructed mesh of the environment.  

    \subsection{Experiment Setup} \label{subsec:exp_set}
\begin{figure}[htbp]
    \centering
    \includegraphics[width=0.8\linewidth]{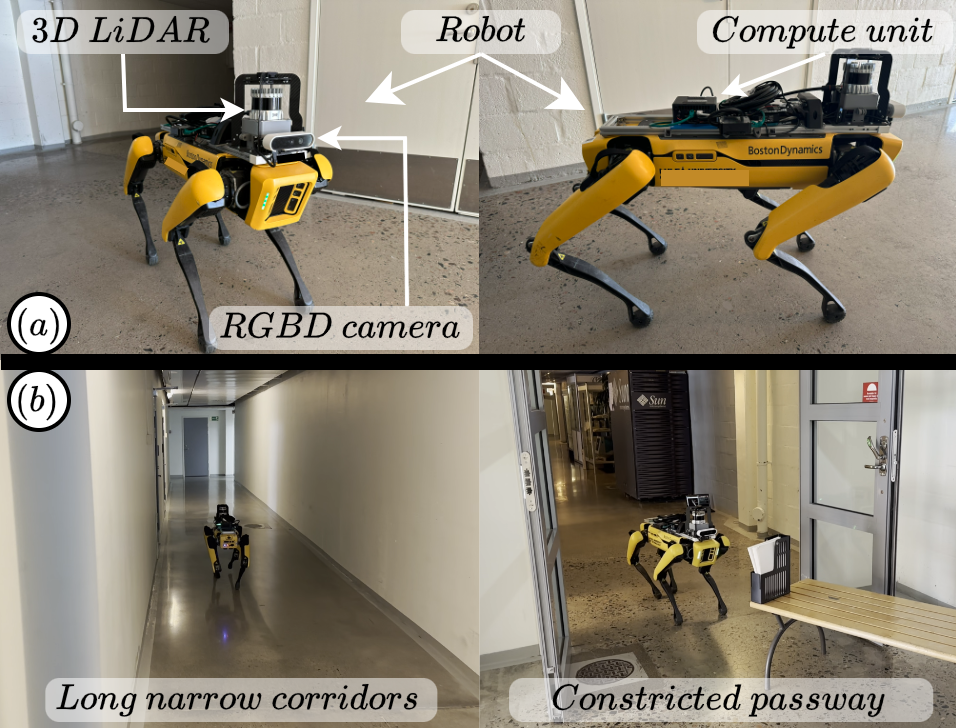}
    \caption{Experimental Setup: (a) Robot with onboard sensors and computer; (b) Inspection environment with narrow corridors and misplaced objects.}
    \label{fig:hardwareSetup}
\end{figure}

\begin{figure}[htbp]
    \centering
    \includegraphics[width=\linewidth]{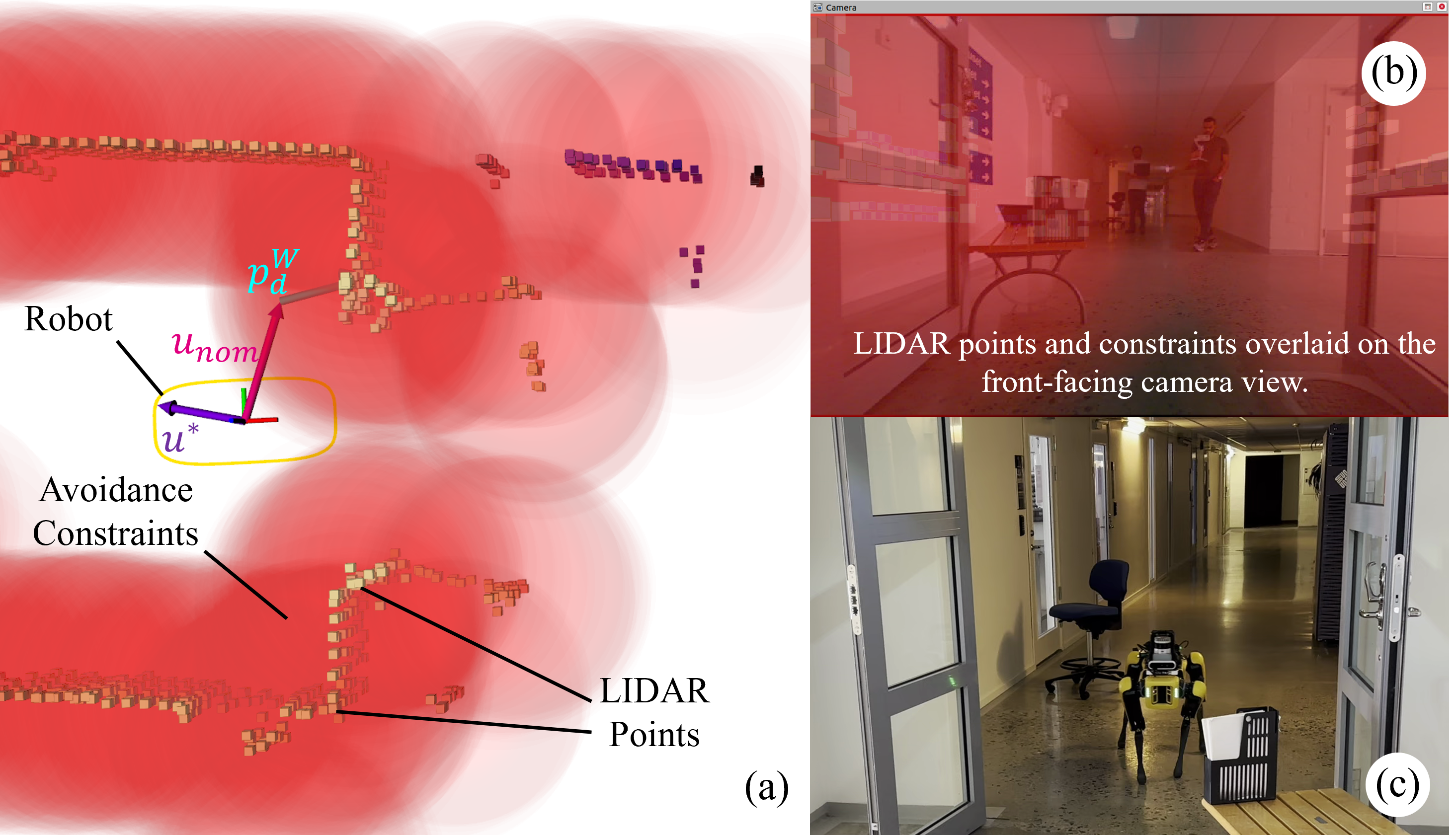}
    \caption{Scenario 1: Conventional distance-based constraints occlude all narrow passages, causing a deadlock. (a) Red circles denote the constraints imposed on each of the acquired LIDAR points (represented as colored cubes). The robot’s body is illustrated as a rounded yellow rectangle scaled to the actual dimensions. The nominal velocity vector $u_{nom}$, and the filtered control velocity $u^*$ vectors are depicted as magenta and purple color arrows, respectively, while the cyan arrow represents the desired position $p^W_d$. (b) The image from the robot's front-view camera is superimposed with LIDAR points (cubes) and the corresponding avoidance constraints in red. (c) External (third-person) view of the robot.}
    \label{fig:narrow_blockage}
\end{figure}

\begin{figure*}[ht]
    \centering
    \includegraphics[width=\linewidth]{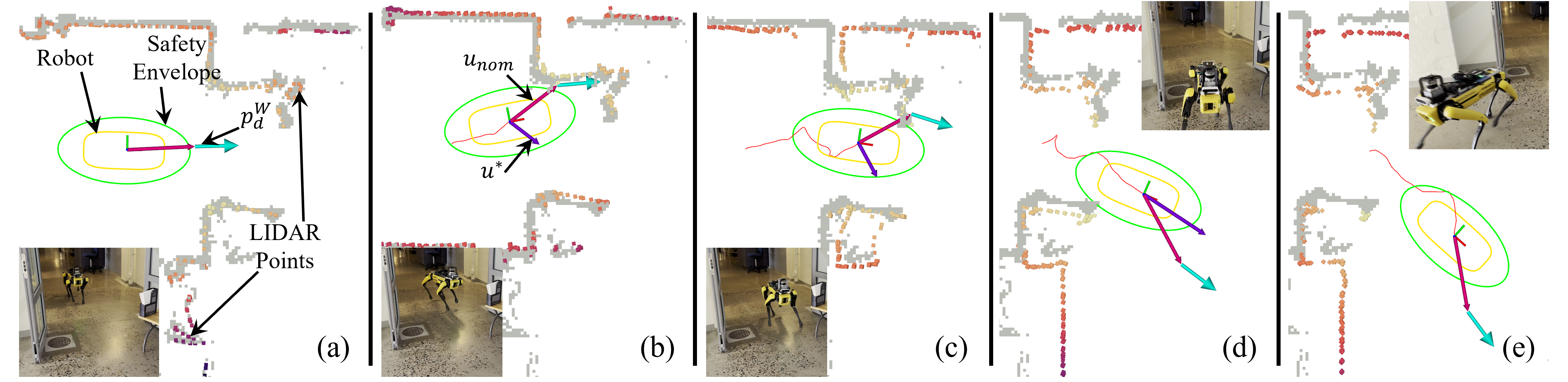}
    \caption{Scenario 1 (Robot-Centric Constraints): The elliptical safety envelope enables the robot to navigate through the narrow passage safely. Note: The red circles in (d) and (e) indicate the drift in the odometry between the static map (gray colored points) and the instantaneous scan (multi-colored points).}
    \label{fig:narrow_pass_seq}
\end{figure*}

The proposed control architecture is evaluated on a visual inspection mission using a Boston Dynamics (BD) Spot quadruped robot equipped with an Orbbec Gemini2XL stereo-camera, a Vectornav VN-100 Inertial Measurement Unit (IMU), and Ouster OS-0 3D LIDAR. We use an Intel NUC embedded computational board running ROS Noetic and Ubuntu 20.04. We use Voxblox~\cite{oleynikova2017voxblox} for 3D mapping. The CBF-based safety filter implemented using the CVXPY library \cite{diamond2016cvxpy} in Python uses all points within a radius of 3.5 m of the instantaneous point cloud from the 3D-LIDAR as safety constraints. The inspection planner is initialized with $\gamma_H=0.5$, $d_{view}=1.5m$, $\alpha=69.4^\circ$, $\beta=45^\circ$. The control parameters are chosen to be, $\gamma = 0.9$, $\kappa = 8.0$, $\alpha = 2.0$, and $K_p = 4\mathbf{I}$ Since the evaluation is conducted over a quadrupedal robot, we ignore the vertical overlap condition throughout the experiments. The results and analysis of these scenarios are presented in the following subsections.

\subsection{Scenario 1: Constricted Passage} \label{subsec:exp_res_1}

In this scenario, we evaluate our proposed robot-centric constraint design compared to a standard point-based distance constraint $(p^W_r - p^W_i)^2 \geq s_d, ~\forall i \in \lbrace 1, 2, ..., N \rbrace$. Here, we chose the safety distance $s_d$ to be 0.8 m (0.25 m more than the quadruped's length) from the obstacles. Figure \ref{fig:narrow_blockage} (a) shows the instantaneous point cloud from the LIDAR and the unsafe region around each point (marked in red). It is noticeable from Figure \ref{fig:narrow_blockage} (a)–(c) that, although there exists a gap between the misplaced obstacle and the door that would, in principle, allow the robot to pass, the designed unsafe regions overlap and thus completely obstruct all available free space. Also, due to faults in planning, the reference is provided in the unsafe region. So, even though the nominal velocity ($u_{nom}$) is pointing towards the desired reference position, the CBF-filter generates a control input ($u^*$) to push the robot away from the obstacles. Therefore, the robot gets into a deadlock position due to conservative modeling of constraints.

In contrast, the robot-centric constraint envelopes the robot's body with an axis-aligned ellipse with $a_x = 0.9 $ m (the major axis is $0.1$ m more than the previous case to provide sufficient safety distance to the robot's corners, which are closer to the ellipse), and $a_y = 0.45$ m. Figure \ref{fig:narrow_pass_seq} presents the sequence of motion taken by the robot to pass through the narrow gap. In Figure \ref{fig:narrow_pass_seq} (a), the nominal and filtered velocities are aligned as the reference position in the open region. But when the robot moves to step (b), the planner provides a reference too close to the opposite door. As the robot approaches the obstacle, the safety filter modifies the velocity input to maintain the safety envelope away from the obstacles. Since the robot reaches the threshold distance for the planner to update its reference, it provides a new reference in (c) behind the obstacle. Again, the robot gets close to the obstacle, but the CBF filter pushes it away into the open region. In step (d), the reference is updated to the other side of the corridor to continue inspection. The CBF minimally invades the nominal control input to move away from the obstacle while moving towards the reference. In step (e), the robot is outside the bottleneck and, therefore, the filtered control input aligns with the nominal control input. Thus, the axis-aligned design helps the robot navigate through the narrow passage while ensuring that the obstacle points stay outside the elliptical envelope around the body.

In Figure \ref{fig:narrow_pass_seq} (d) and (e), the mild drift in the odometry is noticeable, so the current LIDAR points do not align well with the precomputed map. In autonomous missions, these drifts can be hazardous if the robot completely relies on the high-level layers for safety. So, in the next subsection, we evaluate the performance of the controller during a sudden change in the odometry.

    \subsection{Scenario 2: Odometry Failure} \label{subsec:exp_res_2}

\begin{figure}[ht]
    \centering
    \includegraphics[width=\linewidth]{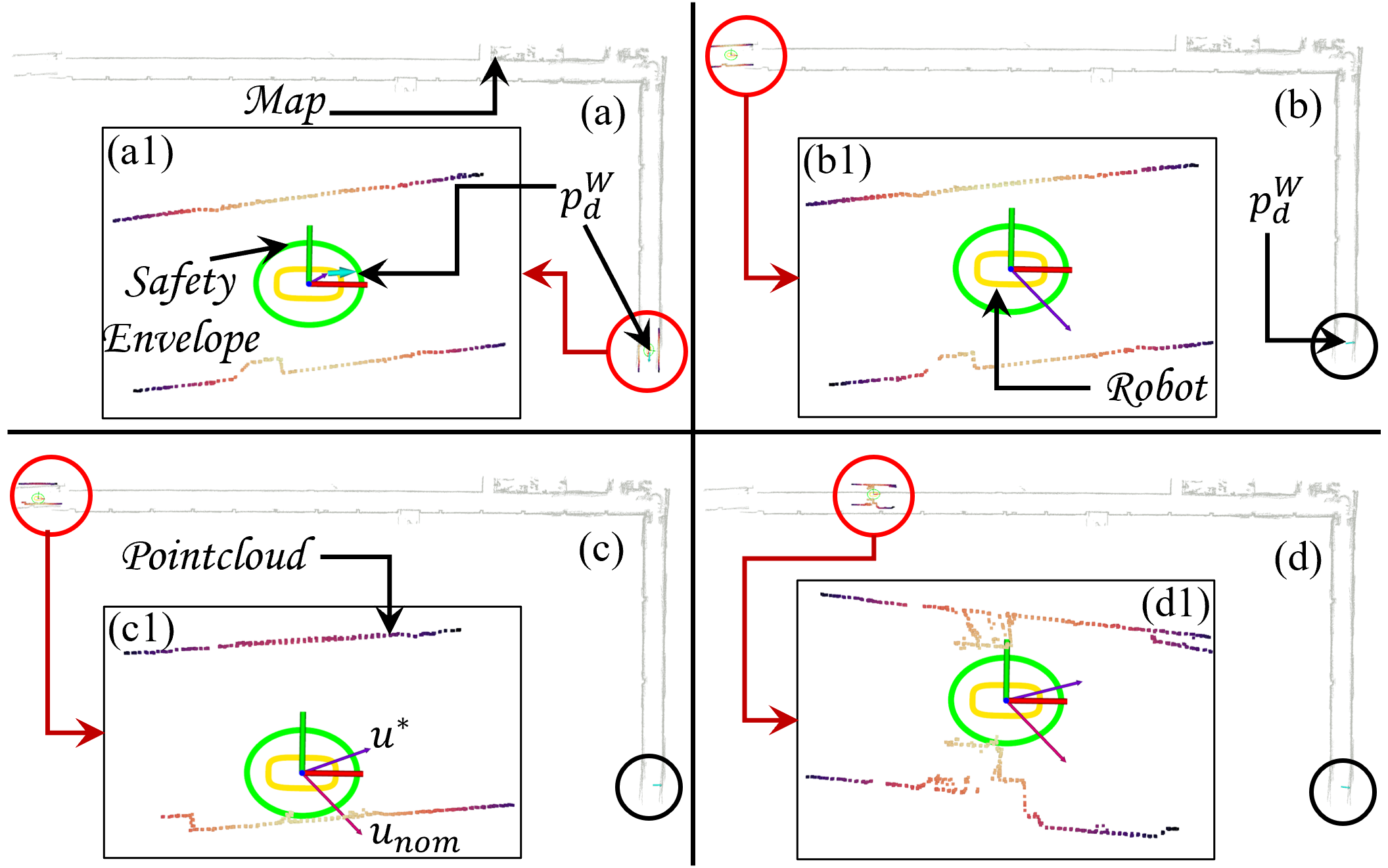}
    \caption{Scenario 2: The odometry jumps between frames (a) and (b). Red circles indicate the robot's position in the map, and black circles indicate the reference position. The magnified images (a1)--(d1) show the LIDAR points in the robot's body frame. (c) robot approaches a wall immediately after odometry drift, but $u^*$ pushes it away; (d) robot safely passes through a doorway chasing the mismatched reference.}
    \label{fig:odom_drift}
\end{figure}
\begin{figure}[ht]
    \centering
    \includegraphics[width=\linewidth]{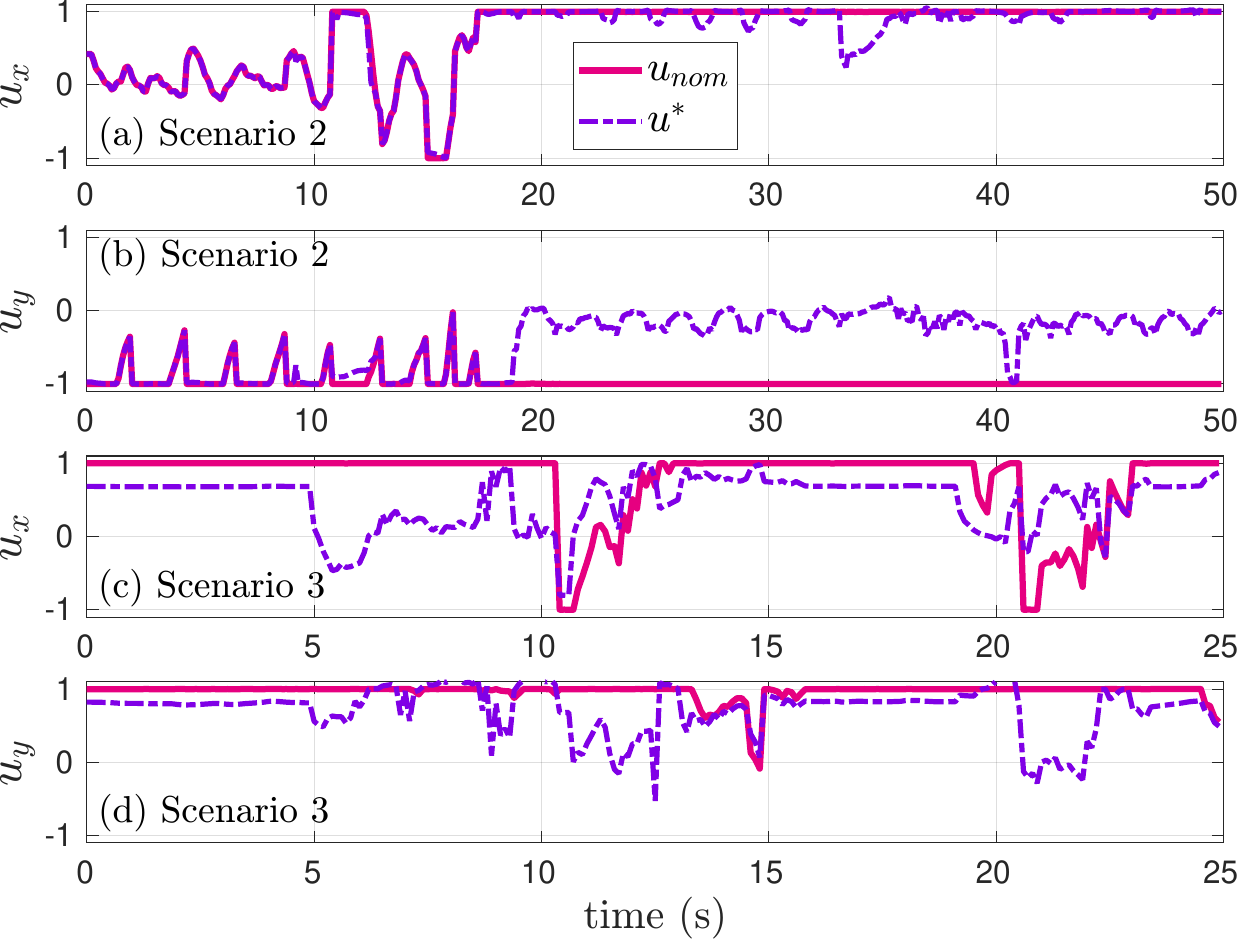}
    \caption{Control velocities for scenarios 2 and 3 generated by the nominal controller and the safety filter.}
    \label{fig:u_values}
\end{figure}

\begin{figure}[ht]
    \centering
    \includegraphics[width=\linewidth]{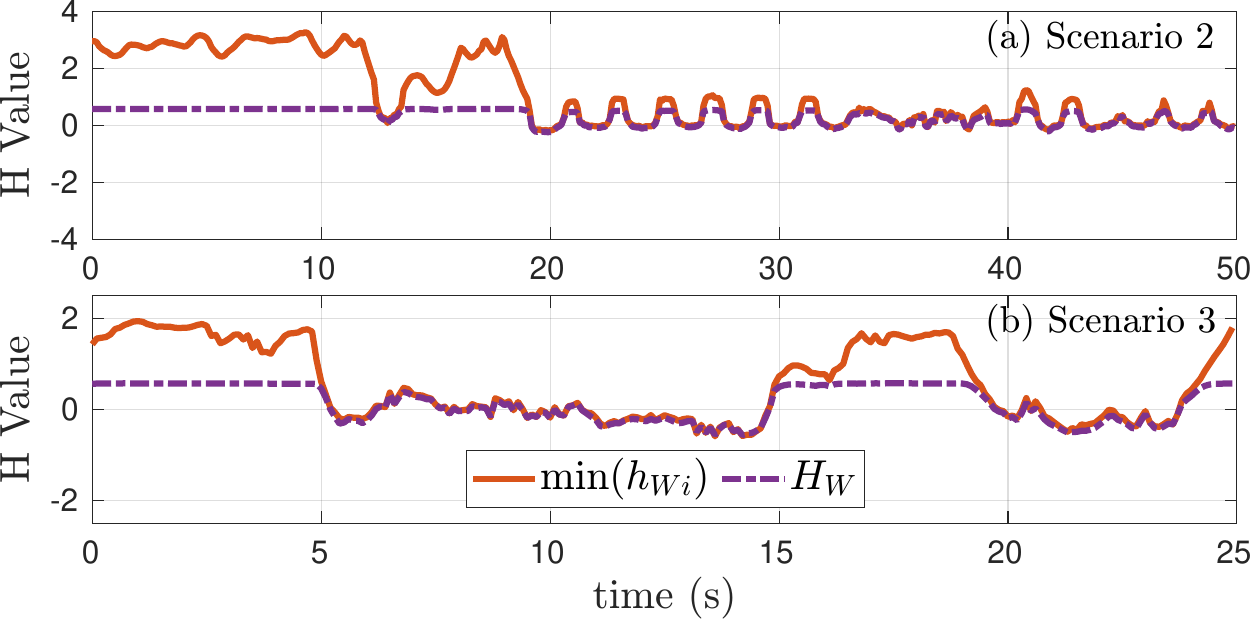}
    \caption{Values of the composite CBF $H_W$ and minimum among the individual CBFs $h_{Wi}$ for scenarios 2 and 3.}
    \label{fig:h_func}
\end{figure}
This subsection presents an evaluation of the controller in the presence of drastic odometric drift occurring due to perceptual aliasing in self-similar structures and incomplete loop closures. In this scenario, the inspection mission is carried out in an underground tunnel consisting of long corridors. Since the corridors are relatively wider $a_y$ is chosen to be 0.6 m for additional safety. Figures~\ref{fig:odom_drift}(a) and (b) visualize two consecutive time instances, between which the robot's position drastically changes (exceeding 100 m) due to similarity between the LIDAR observations caused by repetitive corridor geometry. In Figure~\ref{fig:odom_drift}(b), since the reference position remains unchanged, the nominal controller generates a command that directs the robot to the wall. However, the reactive CBF continues to enforce the safety constraints using the onboard LIDAR measurements, preventing collision and causing the robot to move parallel to the wall while maintaining the safety envelope away from the obstacles. In this process, the robot continues to navigate towards the false goal, while safely passing through a doorway as seen from the LIDAR scans in Figure~\ref{fig:odom_drift}(d).

The jump in odometry is reflected in the velocity control input generated by the nominal controller $u_{nom}$, as seen in Figure~\ref{fig:u_values}(a) and (b). At $16$ s, when the odometry jumps, the nominal control velocity saturates along both axes, attempting to reach the falsely perceived goal. It is also evident that from the beginning of the experiment until the odometry jump, the CBF filter minimally interferes with the nominal control input without any persistent filtering. After the jump (cf. around $17$ s in Figure~\ref{fig:h_func}(a)), as the nominal input drives the robot toward the nearby wall, the value of the composite CBF $H_W$ and the minimum among the values of $h_{Wi}$ transition from a large positive value to a negative value. Consequently, the safety filter modifies the control input to prevent collision with the obstacle. While the CBF filter maintains $H_W$, small boundary oscillations occur when the reference lies inside the unsafe region, which are typical in hardware deployments due to modeling inaccuracies and mechanical limitations.

    \subsection{Scenario 3: Dynamic Obstacles}
\begin{figure}[ht]
    \centering
    \includegraphics[width=\linewidth]{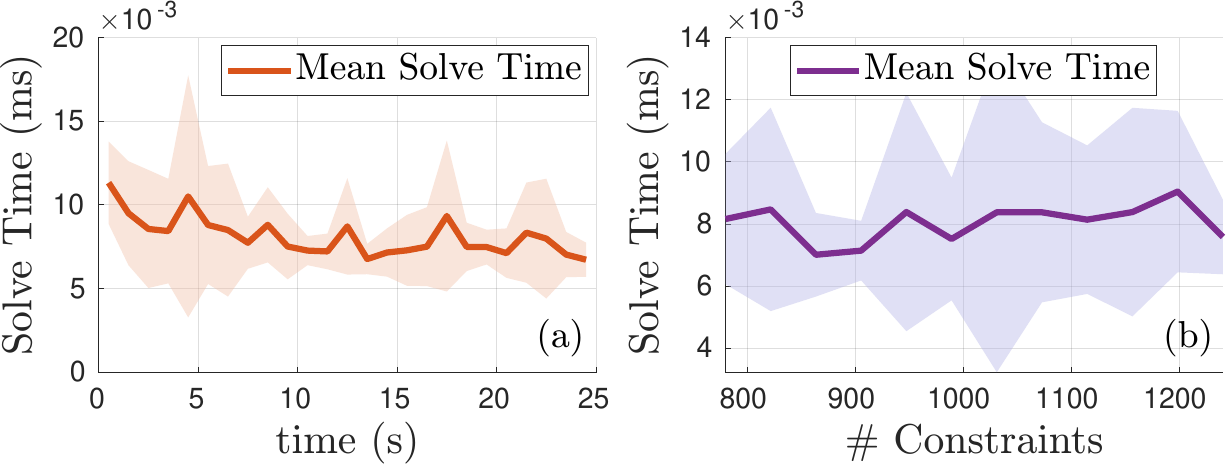}
    \caption{Mean and standard deviation of the solver time plotted against (a) experiment time and (b) number of active constraints.}
    \label{fig:solver_time}
\end{figure}
In this scenario, we introduce dynamic obstacles into the environment by moving a chair and opening a door while the robot performs the inspection task, emulating a busy industrial workspace shared by humans and other robots. Since these obstacles are absent in the precomputed map used for mission planning, the nominal path to the reference position becomes obstructed. We perform experiments in a relatively constricted corridor with static obstacles on both sides of the corridor. So, the CBF filter provides a comparatively smaller input than the nominal input while passing through the corridor, as reflected in the initial period of Figure~\ref{fig:u_values}(c) and (d). At around $9$ s, a chair is moved toward the robot, and at around $20$ s, a door on the side is opened to obstruct the robot's path. The sudden introduction of these obstacles decreases the value of $H_W$, as observed in Figure~\ref{fig:h_func}(b). The filter immediately pushes the robot to keep its safety envelope away from the obstacles by modifying the nominal control input as seen in Figure~\ref{fig:u_values}(c) and (d), where the filtered control input temporarily deviates from the nominal command to steer the robot away from the newly introduced obstacles. Once sufficient clearance is restored, the control input gradually converges back to the nominal value, allowing the robot to continue the inspection task. While unforeseen obstacles are unavoidable in busy industrial environments, the proposed control architecture enforces the safety envelope around the robot in real time, preventing collisions while remaining minimally invasive to the nominal task.

Finally, we analyze the time-criticality of the proposed safety filter. Figure~\ref{fig:solver_time} shows the mean and standard deviation of the solver time measured on the real hardware during Scenario 3. Figure~\ref{fig:solver_time}(a) indicates that the average time required to enforce the safety constraints remains below \textbf{12} microseconds during the entire experiment. Notably, \textbf{99.6 \%} of the solver calls complete within \textbf{20} microseconds, with a single outlier reaching 30 microseconds. We further evaluate the computational performance while handling hundreds of constraints by plotting the solver time against the number of active constraints in Figure~\ref{fig:solver_time}(b). Even when simultaneously enforcing between \textbf{800} and \textbf{1300} constraints, the optimization runs sufficiently fast, validating that the CBF filter can reliably ensure safety at control frequency even with unforeseen obstacles in constricted environments.

\section{Conclusion} \label{sec:conc}

In this article, the safety challenges in autonomous robot navigation in spatially constrained dynamic environments are addressed through a robot-centric constraint formulation using onboard sensing. The proposed formulation uses instantaneous LIDAR point cloud observations that induce time-varying constraints, which are simultaneously enforced via a composite CBF framework while remaining minimally invasive to high-level planning processes. The proposed controller is successfully validated onboard a quadrupedal platform performing an autonomous inspection mission in narrow passages under localization anomalies, unsafe references, and adversarial obstacles.

\bibliographystyle{IEEEtran}
\bibliography{root}
\end{document}